\documentclass[manuscript,screen,nonacm]{acmart}

\graphicspath{{fig/}}
\DeclareGraphicsExtensions{.pdf,.jpeg,.png}

\usepackage{booktabs}

\begin{document}

\title{Computer vision-based analysis of buildings and built environments: A systematic review of current approaches}

\author{Małgorzata B. Starzyńska}
\authornote{Corresponding Author}
\email{m.starzynska@network.rca.ac.uk}
\orcid{0000-0002-9575-2920}

\author{Robin Roussel}
\authornotemark[1]
\additionalaffiliation{\department{Computer Science Research Centre}}
\email{robin.roussel@rca.ac.uk}
\orcid{0000-0001-8875-3688}

\author{Sam Jacoby}
\email{sam.jacoby@rca.ac.uk}
\orcid{0000-0002-9133-5177}

\affiliation{%
    \institution{Laboratory for Design and Machine Learning, Royal College of Art}
    \department{School of Architecture}
    \streetaddress{Kensington Gore, South Kensington} 
    \city{London} 
    \country{United Kingdom}
    \postcode{SW7 2EU}}

\author{Ali Asadipour}
\affiliation{%
    \institution{Computer Science Research Centre, Royal College of Art}
    \department{Research Centres}
    \streetaddress{Hester Road, Battersea} 
    \city{London} 
    \country{United Kingdom}
    \postcode{SW11 4AN}}
\email{ali.asadipour@rca.ac.uk}
\orcid{0000-0003-0159-3090}

\renewcommand{\shortauthors}{Starzyńska and Roussel, et al.}

\begin{abstract}
Analysing 88 sources published from 2011 to 2021, this paper presents a first systematic review of the computer vision-based analysis of buildings and the built environments to assess its value to architectural and urban design studies. Following a multi-stage selection process, the types of algorithms and data sources used are discussed in respect to architectural applications such as a building classification, detail classification, qualitative environmental analysis, building condition survey, and building value estimation. This reveals current research gaps and trends, and highlights two main categories of research aims. First, to use or optimise computer vision methods for architectural image data, which can then help automate time-consuming, labour-intensive, or complex tasks of visual analysis. Second, to explore the methodological benefits of machine learning approaches to investigate new questions about the built environment by finding patterns and relationships between visual, statistical, and qualitative data, which can overcome limitations of conventional manual analysis. The growing body of research offers new methods to architectural and design studies, with the paper identifying future challenges and directions of research. 
\end{abstract}



\acmJournal{CSUR}
\acmYear{2022}

\maketitle

\section{Introduction: Computer vision in built environment studies}

A growing number of disciplines, including architecture, explore data-driven applications in the analyses of new or large digital datasets~\citep{vento_traps_2019}. The use of analog and digital data in architectural practice and theory is well-established in studies of design processes~\citep{demirbas_focus_2003}, buildings, and urban fabrics. Common topics include manufacturing~\citep{beghini_connecting_2014}, design sustainability~\citep{peters_computing_2018}, environmental impact~\citep{jalaei_life_2021}, and morphology~\citep{gil_discovery_2012}. The increased availability of digital proposesdata repositories such as Energy Performance Certificates (EPCs) or property- and planning-related records creates new applications and potential to analyse the built environment. Large scale image data processing and acquisition, in particular, form an emerging area of research in the built environment and studies of its design.

Computer vision methods (including image-based machine learning) applied to buildings as well as larger architectural and urban domains can be grouped into four clusters of research: (i)~landmark and place recognition, (ii)~generative design and modelling, (iii)~remote sensing, and (iv)~the analysis of urban environments. Landmark recognition approaches have been reviewed by T. Chen et al.~\citep{chen_survey_2009} and Bhattacharya and Gavrilova~\citep{bhattacharya_survey_2012}, while Garg et al.~\citep{garg_where_2021} compared visual place recognition methods. New applications of artificial design and urban environment modelling were assessed by Sönmez~\citep{sonmez_review_2018} and Feng et al.~\citep{feng_review_2021}. A review of deep learning applications in remote sensing by Ma et al.~\citep{ma_deep_2019} proposed a taxonomy based on four main tasks: image preprocessing, classification, change detection, and accuracy assessment. Lastly, four reviews investigated the use of street-view imagery for the analysis of urban environments~\citep{biljecki_street_2021,cinnamon_panoramic_2021,he_urban_2021,kang_review_2020}.

Although many of these recent reviews touch on aspects of building recognition, there has been no detailed assessment of computer vision-based applications focusing on architectural analysis across image sources, with an emphasis on usefulness and limitations for architectural studies and practice. The opportunities for architecture, however, are numerous: the growing body of research promises benefits for design optimisation, architectural precedent analysis, and policy making by providing new means of evaluating differences or changes in the built environment. This paper is a first systematic review of the state-of-the-art of computer vision in the analysis of the built environment and in relation to applications at different architectural and urban scales. It compares the research foci, computer vision and machine learning approaches, and data acquisition and curation processes found in recent studies to identify trends and challenges of this often transdisciplinary research as well as future directions and value this might bring to architectural and urban design studies. 

\begin{figure}[htbp]
  \centering
  \begin{minipage}{0.5\textwidth}
        \centering
        \includegraphics[width=0.9\textwidth]{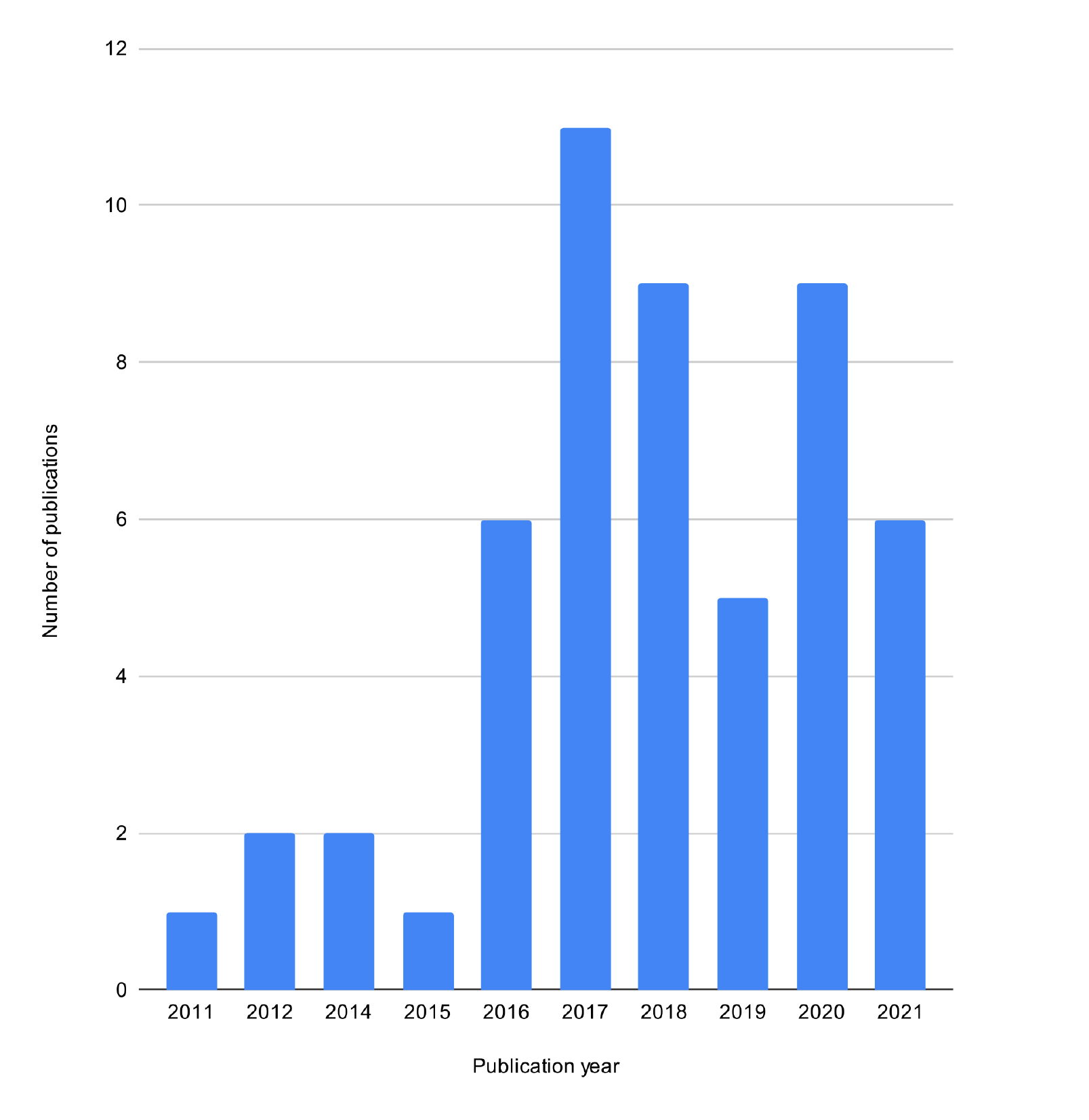} 
        \caption*{(a)}
    \end{minipage}\hfill
    \begin{minipage}{0.5\textwidth}
        \centering
        \includegraphics[width=0.9\textwidth]{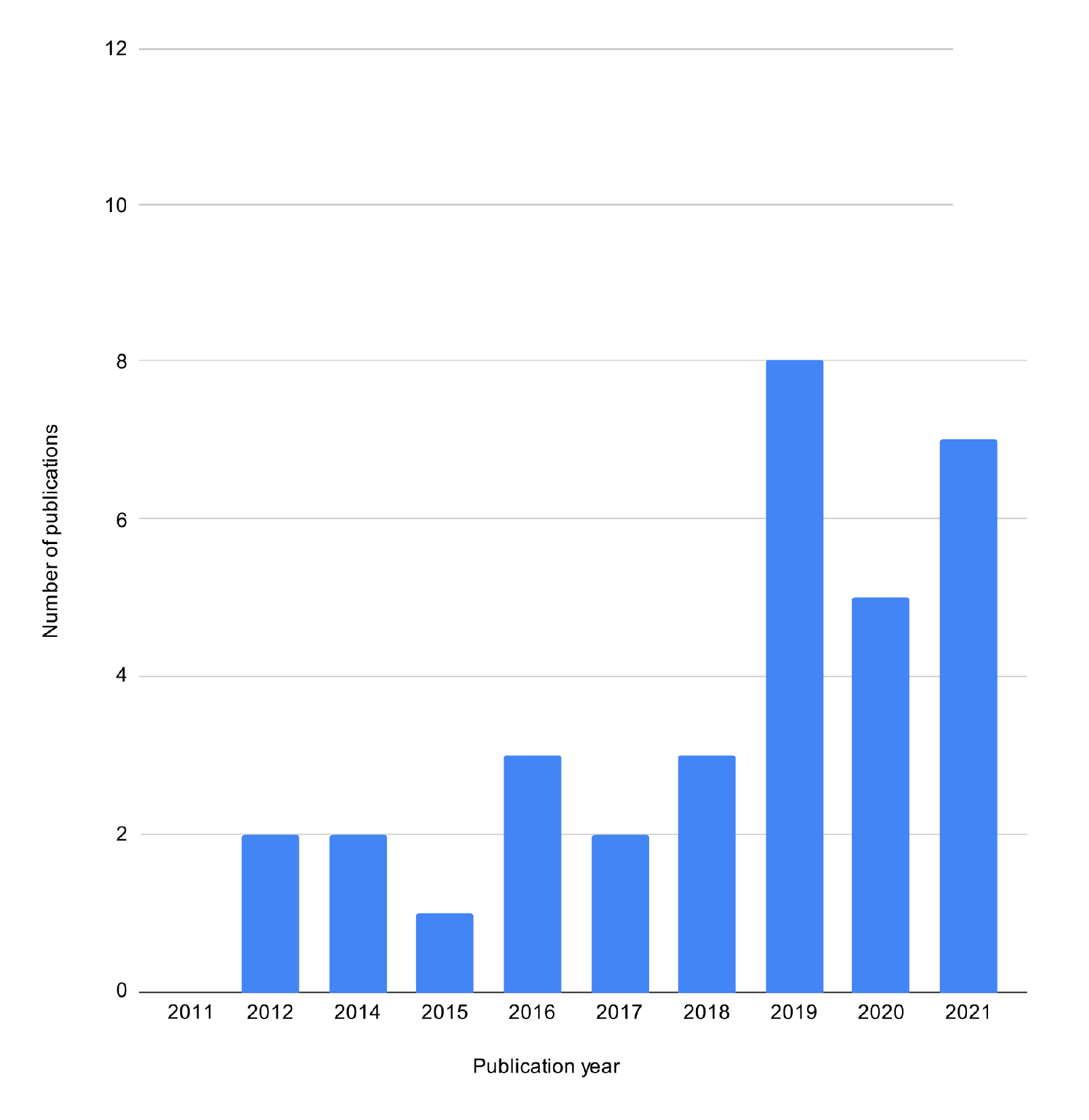} 
        \caption*{(b)}
    \end{minipage}
  \caption{Publications focused on: (a)~innovating machine learning tool by applying them to architectural datasets by year, (b)~formulating novel questions by employing machine learning techniques in the architectural domain by year.}
  \label{fig:architecture}
\end{figure}

A detailed review of 88 sources identified two primary objectives in recent research: 64\% of studies test or improve the performance of existing and novel algorithms by applying them to architectural datasets (Fig.~\ref{fig:architecture}a)  and 36\% assess the methodological benefits and outcomes of using computer vision techniques to ask new questions in the architectural domain (Fig.~\ref{fig:architecture}b). For example, the automation of architectural recognition and classification tasks can expedite otherwise labour- and time-intensive processes such as the recognition of building elements, assigning street views to specific cities or inferring neighbourhood statistics. Past research demonstrates the value of computer vision-based methods of analysis in the architectural domain, such as a correlation of visual and statistical or demographic data. This paper discusses how new questions about urban gentrification, real-estate values or specific characteristics of the built environment can be asked with potential to inform decision-making by designers, occupants, and policymakers. 

The aim is further to contribute to a much needed transdisciplinary evaluation of intersections between computer vision and architectural or urban design studies to strengthen the reliability and methodological rigour of research. Disciplinary differences in understanding and assessing computer vision or spatial design problems can lead to misunderstandings that must be resolved to fully realise the potential of computer vision approaches in architecture.



This review makes the following contributions:
\begin{itemize}
  \item It proposes an application-centred classification of the current literature at the intersection of computer vision and architectural analysis.
  \item It analyses the current trends and research gaps according to that classification.
  \item It assesses the current state of data sources, both in terms of acquisition methods and geographic locations.
  \item It evaluates the reproducibility and comparability of computer vision-driven architectural research, and outlines the main pitfalls found in that field as well as possible solutions.
\end{itemize}

This review is organised into four sections. A methods section provides details on the criteria formation for the inclusion of reviewed papers and sources. This is followed by a summary of the search results and main findings In terms of the contributions highlighted above. A discussion section then  highlights the main trends, challenges and pitfalls faced by researchers in the field. Finally, the concluding section makes recommendations on key future research directions identified through the review.
\section{Methods}
This paper uses the Preferred Reporting Items for Systematic Reviews and Meta-Analyses (PRISMA) method~\citep{page_prisma_2021,page_prisma_2021-1} and a robust multi-stage selection process. Primary and peer-reviewed sources were selected through a keyword search of the IEEE Xplore, JSTOR, Scopus, Semantic Scholar, and Google Scholar databases, as well as via Google Search, to identify studies using computer vision in the context of the built environment. Given a noticeable increase in studies over the last five years, this review is limited to research published from 2011 to 2021 to capture the most recent trends.

\begin{figure}
  \centering
  \includegraphics[width=0.85\textwidth]{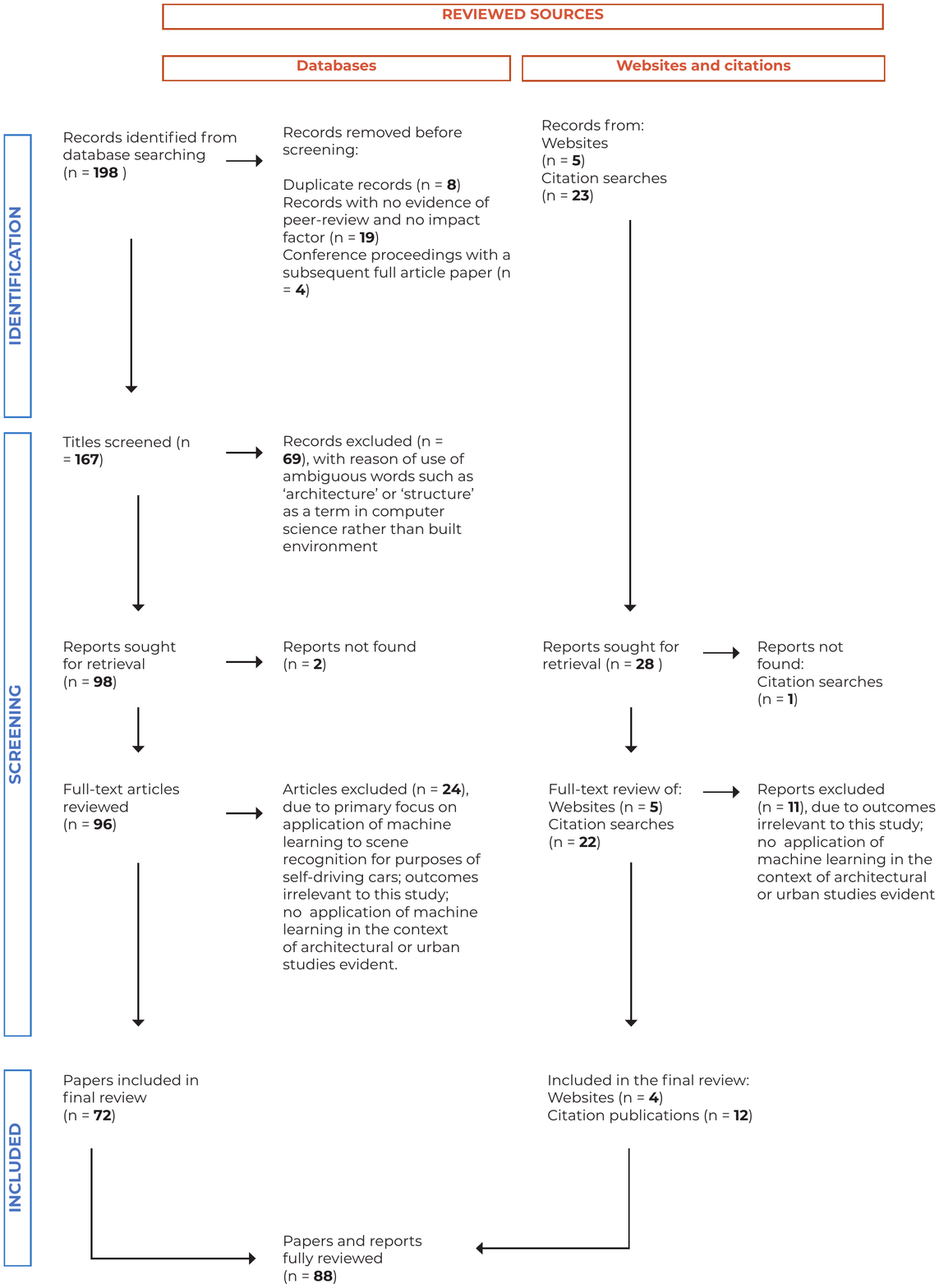}
  \caption{PRISMA flow diagram of source selection for review.}
  \label{fig:PRISMA}
\end{figure}

A challenging aspect of this search is the overlap of terms widely used in both computer science and architecture. The meaning of keywords such as ``architecture'', ``structure'', ``model'', ``design'' or ``building'' depends on disciplinary contexts and might refer to significantly different concepts. For example, the initial search for the keywords ``machine learning'' together with ``architecture'' brought up publications that used the term ``architecture'' to describe the structure of machine learning systems. Using less ambiguous words such as ``façade'', ``urban'', ``city'' or ``ornament'' produced better results in identifying literature relevant to built environment studies. However, it created a risk of excluding papers applying computer vision methods to other aspects of architecture. 
This problem was tackled by using a multi-stage selection process (Fig.~\ref{fig:PRISMA}). Following the keyword search, paper titles, and abstracts were screened in respect to both computer vision and architectural or urban analysis. References in the most relevant papers were also checked. In addition, conference papers (if not published in a journal), online sources (e.g. research project websites), and PhD theses were reviewed and added. This created 226 relevant records. From these, duplicates, papers not peer reviewed or cited in peer-reviewed journals, and conference papers later republished in journals (31 records) as well as papers referring to the ``architecture'' or ``structure'' of computational systems rather than buildings (69 records) were removed, and 3 reports could not be retrieved. This left 123 sources that conformed with the initial search criteria.

A large body of work looked at the application of machine learning in scene recognition for the purpose of navigation and obstacle detection in self-driving vehicles. Although some relate to building analysis, due to their overlap in methodology with papers included in this review, these were omitted from analysis. Furthermore, studies that explored generative rather than analytical systems were excluded, as were those that only used non-visual data or 3D datasets such as point clouds. Both were considered outside the scope of the paper. Removing 35 records, this finally left 88 sources for further analysis in this review.

Two types of information in the publications were compared: (i)~types of algorithms and methods applied in the studies, through the scope of machine learning models and computer vision tasks and (ii)~proposed architectural application (such as urban scene understanding or heritage/style analysis), at different scales (from building detail to satellite) and via different data sources. 
\section{Findings}
\subsection{Search results}
The compared 88 sources included 29 conference papers, 54 journal papers, 5 online reports, and 1 PhD thesis. Most papers were published in computer science journals (38 articles), followed by journals on remote sensing (13 articles) and architectural and urban studies (8 articles). Table 1 details the thematic distribution of journal disciplines per publication.

\begin{table}[]
  \caption{Publication categories of the works analysed in this review.}
  \label{tab:categories}
  \small
  \begin{tabular}{@{}lr@{}}
\toprule
Publication category              & Number of publications          \\ \midrule
Computer science and technology   & 43                              \\
Remote sensing                    & 14                              \\
Architecture and urban studies    & 8                               \\
Geography/geoinformation          & 4                               \\
Computer science and design       & 4                               \\
Computer graphics                 & 3                               \\
Technology                        & 2                               \\
Computer science and architecture & 1                               \\
Economics                         & 1                               \\
VR                                & 1                               \\
Environmental Research            & 1                               \\
Natural Sciences                  & 1                               \\
PhD Thesis                        & 1                               \\
Online sources                    & 4                               \\ \bottomrule
\end{tabular}
\end{table}

\subsection{Computer vision in architectural analysis}
Visual inspections and analysis are standard in building condition surveys and evaluations, with photographs providing a physical record and visual evidence. For example, a visual analysis of façades to determine architectural styles or existing service provisions can be used to establish a building’s age and dwelling type, which might infer typical internal layouts and building maintenance problems. At the same time, building elements such as window or casement types are indicators of thermal performance and used to establish a building’s EPC rating. The size and location of windows can also provide quantitative information about spatial and environmental aspects of quality, such as internal daylight and sunlight penetration. 
Computer vision algorithms can speed and scale up processing visual information in cases such as building condition surveys. Generalising, visual data is currently used in four areas of architectural analysis: building classification, detail classification, qualitative environmental analysis, building condition survey, and building value estimation.

\subsubsection{Image classification of building style and typology}

Identifying and classifying architectural styles and typologies is used in historical or precedent studies in architecture, with the visual analysis of building properties part of a process to identify non-visual attributes. Stylistic features might indicate a property’s age and region or construction type. For example, Victorian buildings in England have a distinctive façade design and internal layout. In addition, building uses can be partially indicated by their façade, with computer vision methods utilised for this task by classifying architectural styles~\citep{lindenthal_machine_2021,mercioni_study_2019,zhao_architectural_2018} and typologies~\citep{kang_building_2018,alhasoun_urban_2019,chen_classification_2021}.
A classification of the whole image is typical for a high-level analysis of architectural features such as overall urban characteristics~\citep{hu_classification_2020} or style or use classification~\citep{li_building_2017}. Chu and Tsai~\citep{chu_visual_2012} exploit a graph-mining algorithm to analyse images for repetitive visual patterns that differ between architectural styles. Obeso et al.~\citep{obeso_architectural_2016} use a CNN to classify Mexican architectural styles, with visual saliency introduced in the algorithm’s network pooling layers to filter relevant features for deeper network layers. Llamas et al.~\citep{llamas_classification_2017} compare the performance of different types of CNNs such as AlexNet or GoogLeNet when trained on pre-labelled images of heritage buildings, and Guo and Li~\citep{guo_research_2017} explore improvements to LeNet-5 when applied to architectural style classification tasks. 

In addition, the website Classify House A.I. allows users to upload an image of a house and, through a computer vision-based analysis, determines which of the 31 architectural style available classifications can be recognised in a building’s exterior (Classify A.I). In another example, Davies~\citep{davies_street_2019} trains an Inception V3 network to recognise Georgian architecture from GSV images of London. Likewise, Alhasoun and González~\citep{alhasoun_urban_2019} use a CNN trained to match GSV images to their corresponding US towns based on a visual classification of urban street contexts or to classify street frontages~\citep{law_street-frontage-net_2020}. Deep-learning models are also used to measure visual similarities between the styles of different architects by Yoshimura et al.~\citep{yoshimura_deep_2019}.

To enhance the architectural benefits of building image classifications, more than one characteristic should be considered at the same time, as many exceptions to ``ground truth'' data can be found across all architectural styles and typologies. For example, dwelling houses might have been converted or changed their use while façades remained the same. Although stylistic and typological features can indicate use and occupancy, they are only one factor, with a more nuanced multi-factor reading needed for reliable estimates. 

\subsubsection{Building detail detection and classification}

Building details, similar to style and typology, can have visual features specific to where a building is located, local climate, and the period it was built in. For example, traditionally high-pitched roofs are found in regions affected by heavy snowfall. The detection and classification of building elements involves prior feature extraction and, like style classifications might use indicators such as window designs~\citep{shalunts_architectural_2011} or face-recognition algorithms (applied to sculpted heads of humans and gargoyles) as a determining feature~\citep{shalunts_detection_2017}. A set of stylistic elements extracted from street-view images is used to determine features typical for Paris (or those untypical) in Doersch et al.~\citep{doersch_what_2012}. In other examples, a bounding-box based object-detection approach separates building details, either extracting whole building façades from the image and then assigning to them a particular style based on their features~\citep{xu_architectural_2014} or extracting façade details to analyse specific building elements~\citep{goel_are_2012,mathias_automatic_2012,zhang_recognizing_2014,llamas_classification_2017,davies_street_2019,wei_computer_2019}. In other cases, semantic segmentation is applied to detect roof typologies and hedgerows maintenance levels from satellite images~\citep{orlowski_uks_2017}, to map green and solar roofs~\citep{wu_roofpedia_2021} or in detail-oriented style analysis where authors train a classifier to distinguish Flemish, Renaissance, Haussmannian, and Neoclassical styles~\citep{mathias_automatic_2012}.

Of the reviewed papers, almost a third (30 records) discuss semantic segmentation, which is key to extracting elements – either whole façades from their urban context~\citep{mathias_atlas_2016,fond_facade_2017,gong_mapping_2018,liu_deepfacade_2020} or façade elements such as doors and windows~\citep{despine_adaptive_2015,mathias_atlas_2016,armagan_semantic_2017,kelly_bigsur_2017,lotte_3d_2018,fond_model-image_2021,liu_deepfacade_2020,zeng_deep_2020,zhong_city-scale_2021}. In research that focuses specifically on semantic segmentation, the extraction of buildings and their elements remains a problem of machine learning techniques and only becomes an architectural question if it is forming part of a larger research process. This includes research on architectural challenges at a scale and complexity difficult to complete using manual methods, such as extracting roof or façade textures to increase the quality of texture patterns in 3D virtual urban models~\citep{despine_adaptive_2015}, reconstructing urban 3D models~\citep{gui_automated_2021, hu_imgtr_2021}, or automating building change detection~\citep{taneja_geometric_2015}.

\subsubsection{Qualitative analysis}
The exploitation of computer vision in qualitative analysis is still in its infancy, yet, has noticeably increased in recent years. The objective is mostly to assess the quality of streetscapes or to establish new links between the aesthetics of an urban environment and statistical data – on education, unemployment, housing, living environment, health, or crime~\citep{arietta_city_2014,gebru_using_2017,suel_measuring_2019}. 
The research project Streetscore~\citep{naik_streetscore_2014} applies Support Vector Regression to predict whether a given streetscape is perceived as safe or unsafe by viewers, and both Dubey et al.~\citep{dubey_deep_2016} and Min et al.~\citep{min_multi-task_2020} study perceptual attributes such as ``safe'', ``lively'', ``boring'', ``wealthy'', ``depressing'', and ``beautiful'' based on GSV images of several US cities. A similar study investigates how visual qualities affect how a street is perceived as walkable~\citep{yin_measuring_2016}. In a similar research, Quercia et al.~\citep{quercia_aesthetic_2014} compared the aesthetic qualities of different areas of London. A crowd-labelled dataset of street-view images from Boston and New York is further used to create perceptual maps for 21 US cities~\citep{li_hierarchical_2017}. The online platform Scenic-Or-Not explores the rating of 200,000 images in relation to perceived qualities of outdoor space and a CNN is applied to analyse and extract key features common to positive scores~\citep{seresinhe_using_2017}. Ilic et al~\citep{ilic_deep_2019} used a siamese CNN to GSV images of properties in Ottawa to determine levels of gentrification. Similarly, visual preferences were examined through semantic segmentation to understand how individual components such as building façades or greenery relate to perceptions of street space quality~\citep{ye_visual_2019}. {\v S}{\'c}epanovi{\'c} et al.~\citep{scepanovic_jane_2021} parsed satellite imagery of six Italian cities to predict urban vitality criteria based on the theories of Jane Jacobs. Neighbourhood vitality was also studied in Wang and Vermeulen~\citep{wang_life_2021}. Lastly, two recent studies use semantic segmentation and k-means clustering in their urban colour analysis~\citep{ding_quantitative_2021,zhong_city-scale_2021}. 
In the analysed studies, qualitative analysis often requires combining several datasets or applying a multi-stage methodology, or both. This has potential for informing design decisions around a building’s form and mass, aesthetics, programme, townscape relationship or user experience, as qualitative assessments are already frequently used in architectural practice. 

\subsubsection{Building condition and value estimation}

Image data is an established means of assessing building conditions and property values, using a qualitative evaluation of building conditions and a quantitative analysis of various building features. Accordingly, several papers look at property price estimation based on a visual assessment. Law et al.~\citep{law_take_2019} use a CNN to automatically extract visual features from GSV images to estimate house prices in London, UK. Lindenthal and Johnson~\citep{lindenthal_machine_2021} combine a traditional hedonic model with architectural style classifications to estimate sales price premia in relation to architectural styles at the building and neighbourhood levels, demonstrating that machine learning classifiers can perform as reliable as human experts in mass appraisals. Wang et al.~\citep{wang_house_2019} explore how an aesthetic value might indicate property prices. Similarly, Poursaeed et al.~\citep{poursaeed_vision-based_2018} estimate house prices based on visual and textural features, with the dataset including interior and exterior images of buildings that are classified according to levels of perceived luxury. In Muhr et al. ~\citep{muhr_towards_2017} satellite images are used to automate assessment of location quality. 
Computer vision algorithms have further assisted in optimising manual tasks of labelling real estate data. Long short-term memory (LSTM) classification algorithms and fully connected neural networks (FCNNs) are applied to real-estate scene classification to automate the labelling of exterior and interior features ranging from types of rooms~\citep{cao_classification_2019} to countertops~\citep{bappy_real_2017}. As some of the image data is of insufficient quality, the studies also use image enhancement processes. 
The visual approach in the condition assessment of buildings tends to focus on image patch analysis. Examples of this include determining single-family house conditions based on building elements such as windows or roofs~\citep{koch_visual_2018}. Zeppelzauer et al.~\citep{zeppelzauer_automatic_2018} automate building age estimations through a two-stage approach, first training a CNN to learn the age characteristics at patch level and, second, globally aggregating patch-wise age estimates of an entire building. In another study, Hoang~\citep{hoang_image_2018} applies SVM to the image analysis of building walls, with particular attention paid to recognising building cracks as indicators of fabric deterioration.
As building condition and value estimation are conventional applications of visual analysis, the discussed studies automate already established processes. A visual building condition survey can assist in estimating property values, forecasting maintenance costs, or assessing a building’s state of repair, including dangerous structural deterioration. Consequently, photographs are often used as evidence, for example, in building surveys or tenancy-related inventories.

\subsection{Trends and gaps of computer vision in architectural analysis}

\begin{figure}
  \centering
  \includegraphics[width=1\textwidth]{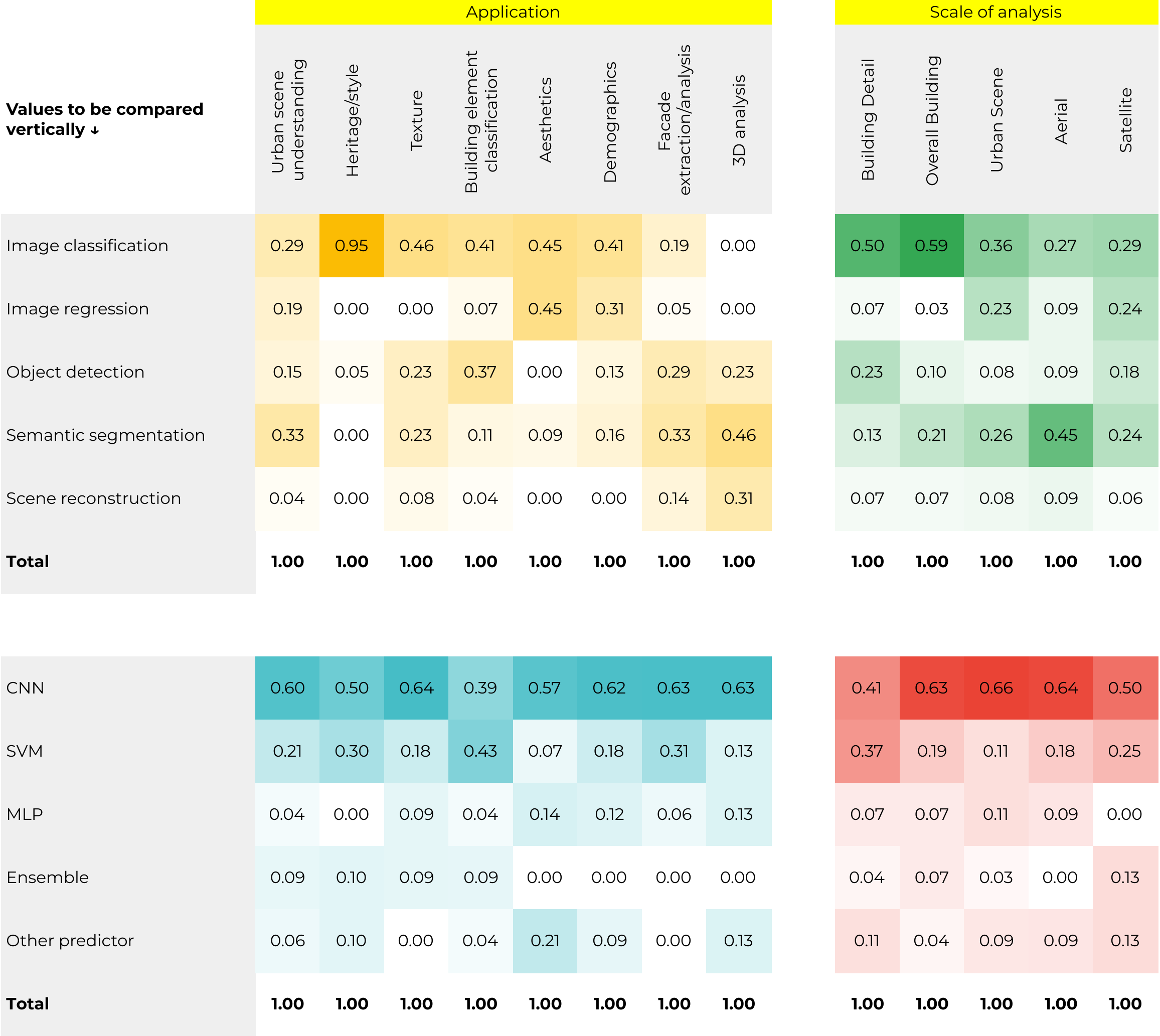}
  \caption{Relative co-occurrences (column-wise). Each value represents the probability of occurrence of a computational method or model given an architectural application or scale.}
  \label{fig:occurrence-cols}
\end{figure}

\begin{figure}
  \centering
  \includegraphics[width=1\textwidth]{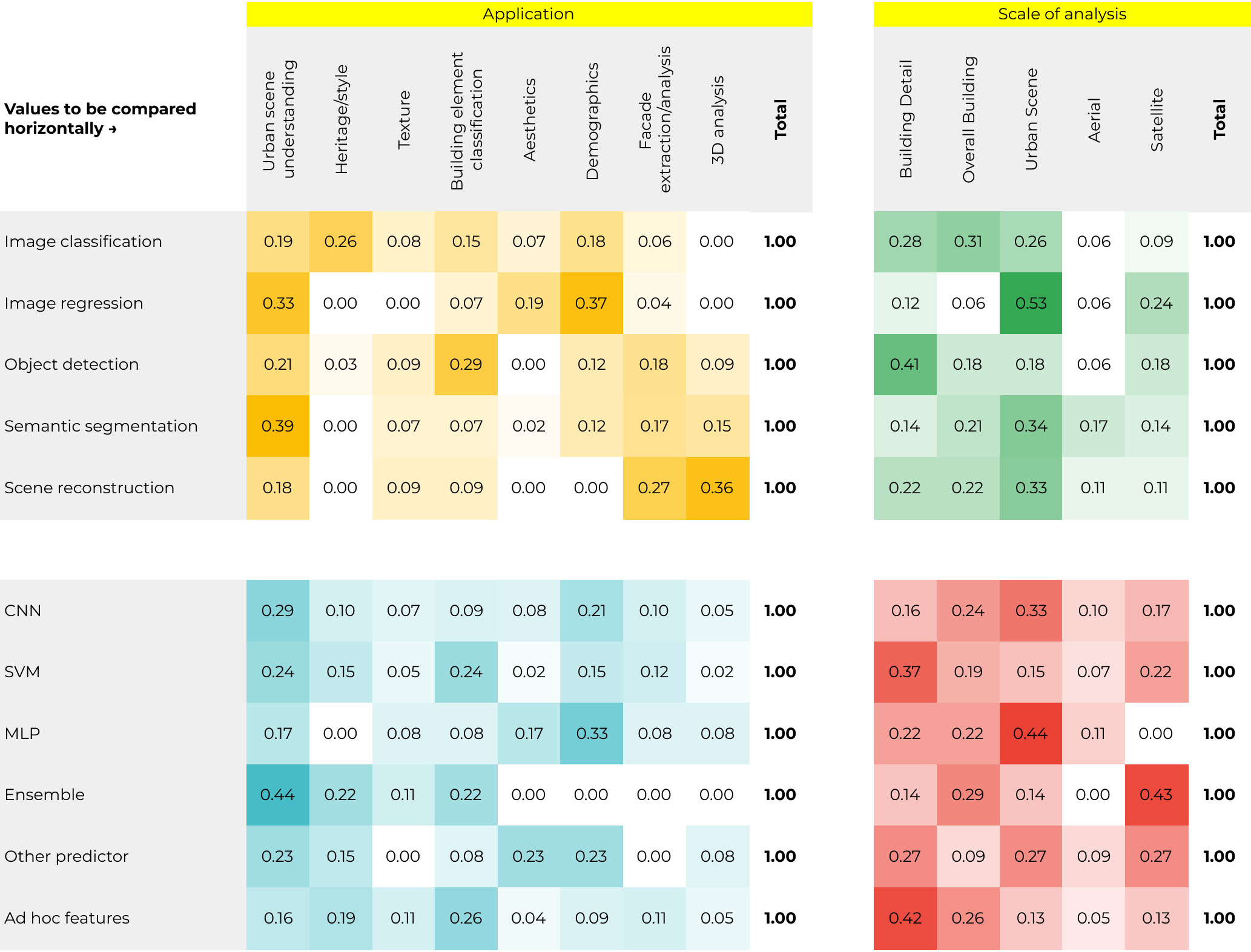}
  \caption{Relative co-occurrences (row-wise). Each value represents the probability of occurrence of an architectural application or scale given a computational method or model.}
  \label{fig:occurrence-rows}
\end{figure}

This section proposes a quantitative analysis of the relationship between architectural applications and scales on the one hand, and computer vision tasks and machine learning methods on the other. This analysis is done in both directions (what is the likelihood of an algorithm given the application, and vice versa?) via relative co-occurrence matrices (Fig.~\ref{fig:occurrence-cols} and~\ref{fig:occurrence-rows}). These matrices differ from typical contingency tables, as categories are not mutually exclusive. Therefore, a single article may be counted several times, and statistical methods such as the chi-squared test cannot be applied to quantify correlations precisely. Moreover, a co-occurrence does not mean that a specific method was used to solve a given problem, only that a method and problem were present in the same publication. But the co-occurrences still give a general sense of the relationships between architectural aspects and computational approaches. While small differences between co-occurrences values are not necessarily meaningful, high contrasts do suggest underlying correlations.

Fig.~\ref{fig:occurrence-cols} shows the relative occurrences of computational approaches for a given architectural application or scale. It reflects which algorithms are most likely to be used for a given application. For instance, it can be seen that CNNs dominate architectural applications and scales, which is consistent with their widespread popularity in other fields. Some applications, such as urban scene understanding, demographics, and façade extraction involve a relatively diverse range of computer vision approaches. Others, such as heritage/style and aesthetic analysis have only been approached from specific angles (image classification for the former, classification and regression for the latter). This may reveal a knowledge gap in terms of how other algorithms and methods could perform on these applications. For instance, what if style analysis was cast as a regression problem along various perceptual dimensions (classic/modern, rural/urban, etc.), rather than a classification problem? Could aesthetics studies benefit from object detection?
In terms of machine learning methods, while CNNs dominate most architectural applications and scales, perhaps the more interesting case is where they do not: the classification of building elements (thus at the scale of building details), where SVMs occur more frequently. One explanation is that such an application typically involves images of façades, often rectified, and thus present a relatively consistent structure. In this specific case, the image invariances learned by deep CNNs in exchange for larger amounts of data might not be as useful.

Analysing the relative occurrences of architectural applications or scales for a given computational approach, as shown in Fig.~\ref{fig:occurrence-rows}, also reveals interesting patterns. While computer vision tasks such as image classification, object detection, and semantic segmentation are distributed across most architectural applications and scales, others appear to occur in more specific contexts. For instance, the scale of image regression is mostly that of urban scenes (i.e., at street or neighbourhood level). Although image regression could theoretically be used at smaller scales, one explanation for this limitation is the difficulty of cross referencing data sources at the building level, since current image datasets usually do not identify individual buildings (except for landmarks). Overcoming this challenge could unlock significant potential for future architectural studies.
Machine learning methods tend to be distributed across all applications and scales, except for ensemble models, although the sample size is too small (n=6) to generalise. Ad hoc features (which include SIFT, Haar, HoG, steerable filters, etc.) are a special case in this table, as they are not a predictor but an intermediate representation that is input into other models (e.g., SVMs). Since they were very popular in computer vision before deep learning methods were developed, they were added to the tables as an interesting indicator of older methods that continue to be useful. In particular, Table~\ref{fig:occurrence-rows} shows a distribution across all architectural applications and scales, with a stronger co-occurrence at small scale. This, as above, can be explained by rectified façade images being more structured and, therefore, allowing models that use smaller amounts of data to perform relatively well.

\begin{figure}
  \centering
  \includegraphics[width=1\textwidth]{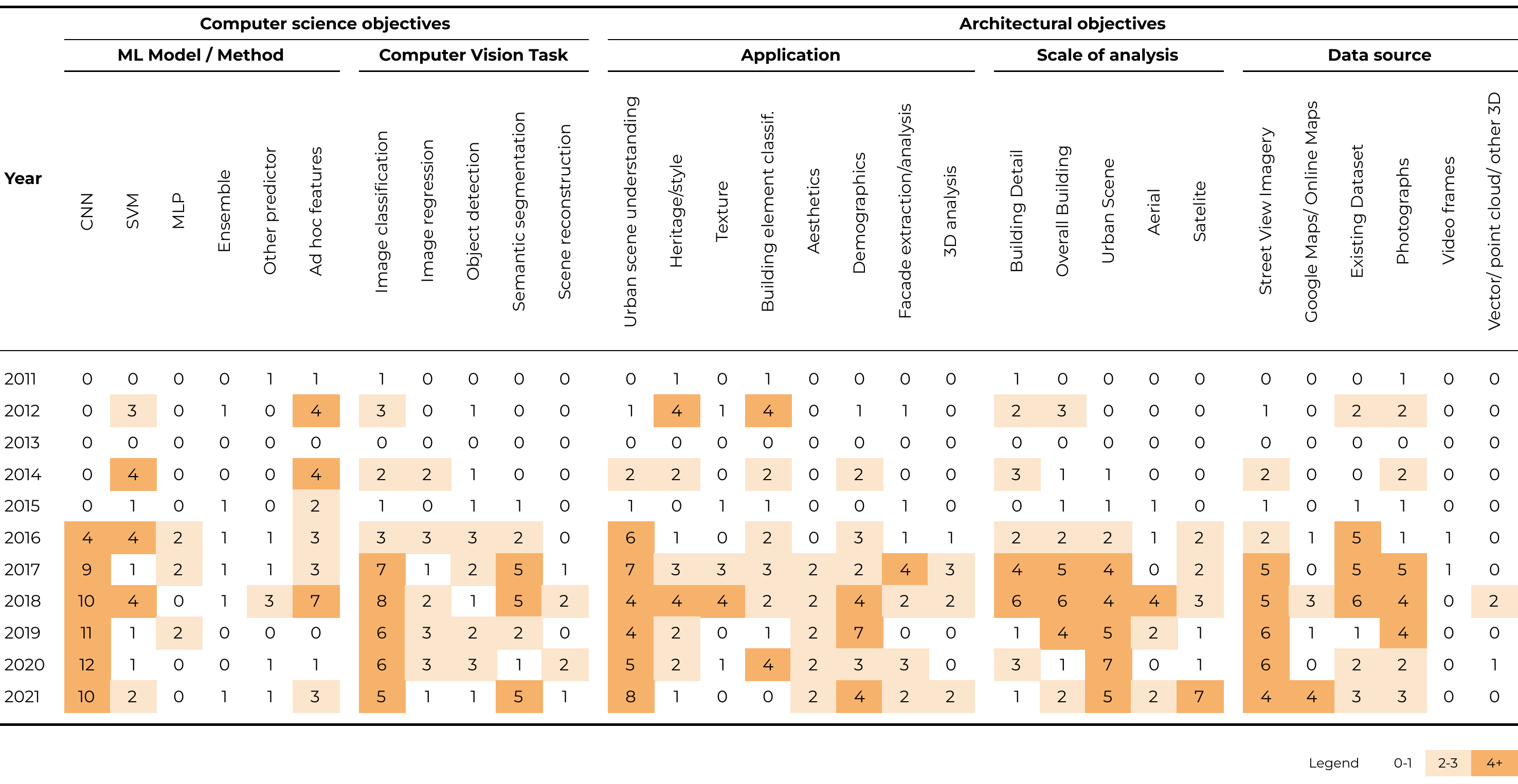}
  \caption{Number of publications per year for each computational and architectural aspect.}
  \label{fig:occurrence-time}
\end{figure}

A temporal analysis can further help to understand relationships highlighted by the previous tables. Fig.~\ref{fig:occurrence-time} breaks down publication numbers for each year across all architectural aspects and computational approaches. In terms of architectural applications, problems such as urban scene understanding, texture, aesthetics, facade, and 3D analysis seem to only adopt computer vision approaches with the development of deep CNNs. A similar relationship can be seen at larger architectural scales (urban scene and above). Other applications, such as heritage, style, and building element analysis were able to benefit from earlier computer vision methods (typically SVMs used with ad hoc features for image classification or object detection). This suggests that the ability of deep CNNs to process large amounts of unstructured data has significantly expanded the applications of computer vision to architectural and urban studies.

\subsection{Data sources and curation}

In this section, data sources are characterised in two ways: first by acquisition method, and second by geographic location. While the former has important technical implications in terms of image scale, spatial and temporal resolution, dataset size, as well as preprocessing and annotations required, the latter directly influences the generalisability of the techniques and findings presented in the literature.

\subsubsection{Acquisition method}

Architectural and urban research uses a wide range of image sources and acquisition methods, including street-view imagery (36\% of studies), photographs either scraped, downloaded or taken by the authors (34\%), existing data from online repositories (32\%), online maps (9\%), and images extracted from vector or 3D data (4\%) and from video frames (3\%). As noted for instance in~\citep{llamas_classification_2017}, there is a lack of image datasets for architectural applications, which means that very often researchers need to build their own: in fact, 84\% of the works surveyed in this review created their own image dataset.

Of the reviewed studies, 25 use photographs from existing datasets, from generic ones such as ImageNet~\citep{gebru_using_2017}, to building-specific ones such as CMP, eTRIMS or Graz50\citep{affara_large_2016,lotte_3d_2018,hu_fast_2020,liu_machine_2017}. This type of dataset can be used for training, but it is also commonly used as an out-of-sample dataset to evaluate the performance of a model. The main advantage of computer vision datasets is that they are readily available for training and evaluation purposes and require little preprocessing. The images are already labelled, and for some facade datasets, already rectified. Combining these datasets, however, may require a step of label homogenisation. The main drawback of this method is that the images come with almost no context or metadata such as geographic coordinates or timestamp, making it impossible to combine with non-visual data sources. Additionally, existing data sets do not currently have a diverse class group that captures a nuanced representation of architectural features and, therefore, can only be helpful for high-level or proof-of-concept studies. 

An alternative acquisition approach is to download images from websites following a keyword search, including images from Flickr~\citep{goel_are_2012,llamas_classification_2017,yoshimura_deep_2019}, Wikimedia~\citep{xu_architectural_2014,guo_research_2017,llamas_classification_2017}, Google Image Search~\citep{chu_visual_2012,goel_are_2012,poursaeed_vision-based_2018} or various real estate websites~\citep{poursaeed_vision-based_2018,zeppelzauer_automatic_2018}. While the first two allow to easily find images under a permissive CC licence, pictures of landmarks and famous buildings vastly outnumber those of more mundane constructions. The quality of the metadata is the most inconsistent of all acquisition methods because it depends on user annotations, resulting in various misclassification errors. Overall, assembling datasets based on images scraped from image websites requires further reviews of user-generated tags to check for accuracy. Real estate websites fare generally better in that regard, owing to their narrower application scope, but their images typically cannot be redistributed~\citep{zeppelzauer_automatic_2018}, and building locations can be purposefully inaccurate for privacy reasons. 
Some image data is captured by the researchers themselves~\citep{shalunts_detection_2017,verma_machine-based_2019}, even including physical synthetic data such as pictures of artificial avian faeces~\citep{lee_contaminated_2020}. This approach creates an opportunity for a highly-collaborative approach to producing custom-made datasets where architectural experts can advise on an appropriate representation of specific architectural features. While this method offers optimal control and consistency, the size of the datasets entirely depends on the resources available to the researchers, which often results in small datasets or highly localised data. As a result, models trained on these images may perform really well on the original dataset but generalise poorly.

Street-view imagery provides data for 32 studies, with images used at different scales: in their entirety (in streetscape analysis), by extracting individual building façades, or by identifying individual building elements (such as doors and windows). While most studies rely on Google Street View, other sources such as Baidu Total View can also be found~\citep{liu_machine_2017,ye_visual_2019,zhong_city-scale_2021}. Some works may use images directly in their panoramic form, or require a rectification step~\citep{affara_large_2016,bochkarev_detecting_2019}. The pros and cons of street-view imagery for urban research have been discussed at length by Cinnamon and Jahiu in a recent review~\citep{cinnamon_panoramic_2021}. To summarise, the main advantages include rapid data collection at a relatively low cost, dense coverage in some areas, relatively precise geographic coordinates, and the possibility of temporal analyses (although panorama locations are inconsistent over time). The main limitations include occlusions and distortions, as well as uneven spatial coverage and frequency of updates.

Satellite and aerial imagery from different images sources are utilised in 13 and 7 papers respectively. Examples of popular freely accessible image data sources for remote sensing applications include the ISPRS 2D Semantic Labeling Contest~\citep{kampffmeyer_semantic_2016,audebert_beyond_2018},images of Beijing and Wuhan collected by ZY3-01 and JL1-07 satellites~\citep{huang_urban_2021} and the Sentinel satellite programme developed by the European Space Agency~\citep{chen_classification_2021,scepanovic_jane_2021}. These sources offer great spatial coverage and precise geographic coordinates, allowing them to be combined with other data sources (as described in the next paragraph). On the other hand, publicly available remote sensing data often comes in low resolution only. While satellite and aerial data available in the public domain can be used for the urban analysis of whole regions, they lack resolution that allows for analysis at a scale lower than a street block. Plan view of individual properties is a valuable representation of the building that can help gauge information about size, layout, or materials; however, they lack precise representation at publicly available resolutions.

Additionally, some works make use of hybrid datasets, combining either different image sources or image data with other data types. Bódis-Szomorú\citep{bodis-szomoru_3d_2018} investigates datasets combining both street-level and aerial images to automate the updating of 3D urban models. J. Kang et al.~\citep{kang_building_2018} supplement remote-sensing data and geographic information with street-level imagery to develop a broader building-use classification based on individual building analysis. Research combining visual with statistical data includes architectural applications of machine learning that might have not previously been possible or evident when using other analytical approaches. For instance, Helber et al.~\citep{helber_multi-scale_2019} propose a multi-scale machine learning approach to analyse aerial and satellite images in conjunction with socio-economic data to predict property-value classes based on image features, while Jean et al.~\citep{jean_combining_2016} pair statistical data of expenditure measurements from the World Bank’s Living Standards Measurement Study with Google Static Maps and satellite imagery from the Night time Lights Time Series to predict poverty levels.Finally, Su at al.~\citep{su_urban_2021} combined high-resolution remote sensing image and statistical data for the purposes of urban scene analysis. Hybrid datasets are a powerful way to overcome the limitations of a single data source, but they require a significant amount of preprocessing to integrate the data.

Overall, greater importance is given to making data processable by computer vision algorithms than to ensure the quality and accuracy of architectural representation in the data. Some studies therefore provide almost no information on their image sources~\citep{affara_large_2016,bappy_real_2017,ibrahim_urban-i_2021}, and the use of public-domain data without verifying if architectural features are identified correctly is prevalent. This includes Flickr or Google data, as image tags and keywords are often provided by non-experts. None of the research uses available digital image libraries tagged by architectural experts, such as RIBApix (image repository of architectural assets curated by the Royal Institute of British Architects) or the Cities and Buildings Database by the University of Washington. This demonstrates that a more transdisciplinary collaboration would be beneficial to make the most out of the sources available.

\subsubsection{Geographic location}

\begin{table}[]
  \caption{Occurrence of geographic locations in each custom image dataset. (Hong Kong was considered separately from China because of its specific context in terms of architecture and data access.)}
  \label{tab:geolocation}
  \footnotesize
  \begin{tabular}{@{}llr@{}}
\toprule
Location                     & Continent          & Occurrences          \\ \midrule
US                           & North America      & 22                   \\
UK                           & Europe             & 17                   \\
France                       & Europe             & 12                   \\
China                        & Asia               & 9                    \\
Hong Kong                    & Asia               & 6                    \\
Canada                       & North America      & 5                    \\
Germany                      & Europe             & 5                    \\
Netherlands                  & Europe             & 5                    \\
Spain                        & Europe             & 5                    \\
Austria                      & Europe             & 4                    \\
Italy                        & Europe             & 4                    \\
Japan                        & Asia               & 4                    \\
South Korea                  & Asia               & 3                    \\
Switzerland                  & Europe             & 3                    \\
Australia                    & Oceania            & 2                    \\
Belgium                      & Europe             & 2                    \\
Czech Republic               & Europe             & 2                    \\
Denmark                      & Europe             & 2                    \\
Malawi                       & Africa             & 2                    \\
Mexico                       & South America      & 2                    \\
Nigeria                      & Africa             & 2                    \\
Russia                       & Europe             & 2                    \\
Rwanda                       & Africa             & 2                    \\
Singapore                    & Asia               & 2                    \\
Tanzania                     & Africa             & 2                    \\
Uganda                       & Africa             & 2                    \\
Angola                       & Africa             & 1                    \\
Argentina                    & South America      & 1                    \\
Benin                        & Africa             & 1                    \\
Brazil                       & South America      & 1                    \\
Burkina Faso                 & Africa             & 1                    \\
Cameroon                     & Africa             & 1                    \\
Côte d’Ivoire                & Africa             & 1                    \\
Democratic Republic of Congo & Africa             & 1                    \\
Ethiopia                     & Africa             & 1                    \\
Ghana                        & Africa             & 1                    \\
Greece                       & Europe             & 1                    \\
Guinea                       & Africa             & 1                    \\
India                        & Asia               & 1                    \\
Kenya                        & Africa             & 1                    \\
Lesotho                      & Africa             & 1                    \\
Luxembourg                   & Europe             & 1                    \\
Mali                         & Africa             & 1                    \\
Mozambique                   & Africa             & 1                    \\
New Zealand                  & Oceania            & 1                    \\
Romania                      & Europe             & 1                    \\
Senegal                      & Africa             & 1                    \\
Sierra Leone                 & Africa             & 1                    \\
South Africa                 & Africa             & 1                    \\
Sweden                       & Europe             & 1                    \\
Thailand                     & Asia               & 1                    \\
Togo                         & Africa             & 1                    \\
Turkey                       & Europe             & 1                    \\
Ukraine                      & Europe             & 1                    \\
Vietnam                      & Asia               & 1                    \\ \bottomrule
\end{tabular}
\end{table}

Context is essential to understand buildings from an architectural standpoint, whether the goal is to assess building style, age, use, or to combine visual and statistical data. One of the most important pieces of information is the geographic location of the building. In the works surveyed, this information is part of the custom datasets at a high level, most often providing the city or the country to which each building belongs, and much more rarely the exact longitude and latitude~\citep{zhu_large_2020}. No location information is provided in 21\% of the works. For those that do provide the information, the location shows a significant concentration in North American, West European and East Asian countries, as shown in Table~\ref{tab:geolocation}. Moreover, across all countries, images appear to be almost exclusively collected in university cities (with the exception of aerial and satellite imagery).

This bias has strong implications in terms of the generalisability of the results found in the literature. For instance, Lotte et al.~\citep{lotte_3d_2018} observe the poor performance of their model trained on mostly European datasets when applied to a Brazilian dataset. Limited model transferability may be due to different urban characteristics~\citep{chen_mapping_2021} or different architectural styles~\citep{mathias_atlas_2016,kelly_bigsur_2017} across cities and countries. Sometimes assumptions are made which would not generalise across countries: for instance, Nguyen et al.~\citep{nguyen_using_2020} consider ``visible utility wires'' to be an indicator of physical disorder in the USA, while most of Japan’s power grid is above ground. The few papers that sample the long tail of less-surveyed countries focus on issues of economic development and well-being~\citep{jean_combining_2016,yeh_using_2020}, and do not adopt a global focus either.

\subsection{Reproducibility and comparability}

The papers were also analysed in terms of the ability to reproduce their results independently, as well as the comparability of different approaches to the same problem. The term ``reproducibility'' used in this review follows the definition provided by the USA’s National Information Standards Organization as the ability to regenerate ``computational results using the author-created research objects, methods, code, and conditions of analysis'' (NISO RP-31-2021). This is also the definition followed by the Association for Computing Machinery: ``For computational experiments, [...] an independent group can obtain the same result using the author’s own artifacts'' (ACM 2020). Although reproducibility is a weak form of replicability, which refers to the ability to obtain the same results using independently developed artefacts (including code and data), it encourages the clarity and transparency required for replicability. In this review, reproducibility was assessed in first approximation by checking whether the code, custom data and trained models used by the authors are available. Weaker criteria of reproducibility, specific to machine learning methods, were also considered: whether the hyperparameters, data splits (training/validation/test) and ML software were specified, as well as useful information such as training and/or inference times and hardware specifications. The latter two are not strictly necessary for reproduction, but are very useful to budget a deep learning experiment.

The reproducibility criteria are shown in Table~\ref{tab:reprod}. Each criterion is shown as a percentage of relevant works. For instance, all works rely on code, so there are 88 relevant works for ``Code'' and ``Hardware'', but 6 do not use machine learning~\citep{li_mapping_2019,lu_using_2019,quercia_aesthetic_2014,szczesniak_method_2021,xiao_building_2012,zhu_interactive_2020}, so there are only 82 for ``ML software used'' and ``Training/inference time''. The first three criteria (``Code'', ``Custom data'', ``Trained model'') are counted as ``Fully disclosed'' if the artefacts are either directly available online, available upon request, or can be recovered by running a script (for ``Custom data'').

\begin{table}[]
  \caption{Disclosure/availability of each reproducibility criterion, as a percentage of works to which the criterion applies.}
  \label{tab:reprod}
  \small
  \begin{tabular}{@{}lrrrr@{}}
\toprule
Reproducibility criterion & Relevant works & Fully disclosed (\%) & Partially disclosed (\%) & Undisclosed (\%) \\ \midrule
Code                               & 88                      & 13.6                          & 8.0                               & 78.4                      \\
Custom data                        & 74                      & 14.6                          & 4.4                               & 81.1                      \\
Trained model                      & 78                      & 5.1                           & 3.8                               & 91.0                      \\
Hyperparameters                    & 78                      & 39.7                          & 21.8                              & 38.5                      \\
Data split                         & 78                      & 79.5                          & 6.4                               & 14.1                      \\
ML software used                   & 82                      & 65.9                          & N/A                               & 34.1                      \\
Training/inference time            & 82                      & 20.7                          & 6.1                               & 73.2                      \\
Hardware                           & 88                      & 36.4                          & N/A                               & 63.6                      \\ \bottomrule
\end{tabular}
\end{table}

The results presented in Table~\ref{tab:reprod} are stark: 78\% of works are published without any code, 91\% without trained models (for those that train their own), and 81\% of custom datasets are unavailable (including data that used to be available online but has not been maintained). The picture is slightly brighter for ML implementation details: 40\% of works provide a full list of hyperparameter values, 80\% provide their data splitting strategy (training/validation/test, k-fold cross-validation, or a mixture of both), and 66\% disclose which software or framework was used. (It is worth noting, however, that the exact version of the software is seldom mentioned, which is problematic when the hyperparameters are said to be left to ``default values''.) Training/inference times and hardware specifications are also rarely mentioned.

\begin{table}[]
  \caption{Occurrence of each evaluation metric.}
  \label{tab:metrics}
  \small
  \begin{tabular}{@{}llr@{}}
\toprule
Evaluation metric          & Task           & Occurrence \\ \midrule
Accuracy                   & Classification & 46         \\
Confusion                  & Classification & 19         \\
F1                         & Classification & 18         \\
Recall                     & Classification & 15         \\
Precision                  & Classification & 17         \\
Kappa                      & Classification & 4          \\
ROC                        & Classification & 3          \\
Error rate                 & Classification & 3          \\
R²                         & Regression     & 7          \\
MSE                        & Regression     & 5          \\
Average Precision          & Detection      & 5          \\
Kendall's Rank Correlation & Ranking        & 2          \\ \bottomrule
\end{tabular}
\end{table}

The comparability of the works was assessed in terms of the diversity of evaluation metrics used for the same type of task. Table~\ref{tab:metrics} shows the number of occurrences of each metric that appeared in more than one paper. For brevity, similar metrics were gathered under the same umbrella term: accuracy, for instance, can be averaged over classes, over random splits, over time, or computed directly over the whole testing set. Similarly in terms of tasks, ``Classification'' refers to both image classification and semantic segmentation, which means that some metrics can be averaged over images or pixels. Nonetheless, Table~\ref{tab:metrics} already shows quite a wide diversity of metrics for classification, whereas image regression, object detection and image ranking appear to have more consistent metrics. Moreover, many papers use arbitrary combinations of these metrics, rarely justifying why some metrics might be more appropriate than others. It is also notable that only 7 papers provide some measure of standard deviation or variance for their metrics, even though the diversity of ways to compute the mean might prevent comparisons anyway.

\section{Discussion}

\subsection{Research trends}

There is an extensive application of machine learning to the analysis of architectural features in computer science literature. Two significant ways in which recent studies engage with architectural questions and problems can be identified. 
The first optimises algorithmic methods of image analysis by applying them to image data of architectural or urban environments. This uses both existing and custom-made image recognition models. The primary objective of this research is to improve process expediency~\citep{obeso_introduction_2018}, optimise processing tasks, or enhance accuracy~\citep{cohen_rapid_2016,kampffmeyer_semantic_2016}. Once automated, the virtual scene understanding is then deployed in space navigation and virtual visual servoing~\citep{wei_computer_2019}. The contribution of this kind of research is automating work that is otherwise labour-intensive, therefore enabling it to be undertaken faster and at a greater scale and frequency. The shortfall of efficiency-oriented research is that architectural objectives lack sufficient accuracy, and their focus is high-level, which can make the results prone to bias.

The second type of research applies existing or custom-made machine learning systems to the analysis of data useful to answering questions arising in the architectural domain, such as a correlation between streetscape aesthetics~\citep{quercia_aesthetic_2014}, poverty levels~\citep{jean_combining_2016} or visual indicators of architectural styles~\citep{shalunts_architectural_2011,doersch_what_2012,goel_are_2012,obeso_architectural_2016,shalunts_detection_2017,yi_house_2020,lindenthal_machine_2021}. This includes the qualitative analysis of visual clues and perceived attributes that, for example, might indicate urban gentrification~\citep{ilic_deep_2019} or specific urban environmental conditions~\citep{naik_streetscore_2014,quercia_aesthetic_2014,dubey_deep_2016}. But it also includes the quantitative analysis of buildings and their elements~\citep{ding_quantitative_2021}, and is sometimes combined with statistical data. Machine learning provides thereby new methods of analysis that can lead to novel understandings of the built environment and related data. 

The emerging research in the architectural application and computer vision methods is exploratory, and no single dominant interdisciplinary practice strain has yet emerged. This results in a lack of standardised evaluation methods as there are limited or isolated application examples. There is, however, an opportunity to further explore research objectives concerning both problem areas by applying a more collaborative approach between the two disciplines.

\subsection{The importance of datasets}

Image and hybrid datasets play a key role in the training and testing of computer vision models. Images of buildings are studied from different viewpoints and at different scales, such as street level and elevational or satellite and aerial views, and have a wide range of sources: new photographs, ready-made datasets (with various levels of access and permitted use), scraped data from websites, or collected through more or less public APIs (such as GSV panoramas). Almost none of the reviewed records used the same dataset and data preprocessing and curation needs vary greatly. Moreover, very few of these custom datasets are available online, which may be in part due to the unclear legal ramifications it entails (copyright law, breach of End User Licence Agreement, etc.).

There are thus challenges surrounding data acquisition and annotation. Crowdsourcing, for instance, can be unreliable: a level of expertise is required to label and class some images, for example, in an architectural dataset~\citep{ye_visual_2019}. Additionally, data that seems homogeneous might in fact cover different categories. For example, Chu and Tsai~\citep{chu_visual_2012} classify architecture according to Gothic, Georgian, Korean, and Islamic styles – while Gothic and Georgian refer to specific art-historical styles and periods, Korean and Islamic are much broader. More generally, the positionality of the annotators is almost never questioned, even though it might affect qualitative evaluations, such as the definition of “formal” versus “informal” settlements~\citep{ibrahim_urban-i_2021}, or the collection  of “ugly” rooms online~\citep{poursaeed_vision-based_2018}.

Current research shows that architectural image datasets are predominantly utilised in computer science studies that focus on computational issues of image analysis and processing and a single database or building scale. From a built environment perspective, however, there is greater value in emphasising urban and architectural analysis. But this requires overcoming the current limitation of image datasets that lack context and metadata to enable interoperability with other data sources (e.g., statistical surveys). 

\subsection{Pitfalls of computer vision in architectural analysis}

The lack of reproducibility and comparability described in Section 3.5 supports the argument that current research at the intersection of computer vision and architecture is still very exploratory and has not yet focused on making incremental improvements to previous results. While it has been shown that available datasets are lacking from an architectural point of view, the computer vision side could also be improved by providing more transparent implementations and principled evaluation methods. There are additional pitfalls, however, that are particularly prevalent when attempting to tackle architectural issues with a computer vision-based approach. This section describes the three main pitfalls and attempts to provide tips to avoid them.

The first pitfall is ambiguous training data, which is mentioned in 11 of the works that use image classification. Architecture is rife with typologies and classifications (housing type, use class, style, etc.) involving complex definitions and edge cases that attempt to capture the diversity of the built environment. Architectural styles, for instance, are difficult to define precisely~\citep{yi_house_2020}, involve a variety of visual and structural cues, can be mixed together, undergo revivals, and present significant variation across buildings of different use classes. The first step to address this issue is to ensure that the samples are correctly labelled (notably in the case of crowdsourced data). Second, if the error on the training step is unexpectedly high, this might be a case of underfitting: the model does not manage to capture the complex visual relationships that define a class, which might call for a deeper network or a more advanced optimization algorithm. Third, visual data might simply not be enough to separate the classes unambiguously. In that case, a hybrid dataset should be considered.

The second pitfall is class imbalance, especially common in semantic segmentation and object detection (mentioned in 10 works). It typically occurs in urban scene understanding at aerial or satellite scale, where buildings might be quite small compared to their surroundings, or land use classes might be unevenly distributed. This problem can be tackled by various strategies, such as minority over-sampling~\citep{fang_reading_2020}, online hard example mining~\citep{shrivastava_training_2016}, or by modifying the optimisation using e.g. a focal loss~\citep{lin_focal_2017}.

The last pitfall is view sensitivity, which is typical of street-level analysis. Kim et al.~\citep{kim_decoding_2021} provided a thorough quantitative analysis of the impact of panorama locations on attributes computed from semantic segmentation. However, view sensitivity affects all computer vision tasks, due to occlusions, self-occlusions, reflections (for glass buildings) and distortions (for high buildings in narrow streets). While this problem is still open, potential approaches include using a trained model to detect and remove occluded views, combining visual and non-visual data, or combining inferences from different views of the same building.
\section{Conclusion: Challenges and directions of built environment research}

\begin{table}[]
  \caption{Current research summary and future directions}
  \label{tab:summary}
  \small
  \begin{tabular}{@{}lll@{}}
\cmidrule(r){1-2}
\multicolumn{2}{@{}l@{}}{Built environment research} &  \\ \cmidrule(r){1-2}
\multicolumn{1}{@{}l|}{Objectives and applications} & \begin{tabular}[c]{@{}l@{}}Analyse visual-spatial characteristics of building and urban elements for their classification\\and identification (e.g. styles, typologies, geolocations)\\ Rationalise spatial qualities or quantities by rating visual indicators (e.g. building valuations,\\building conditions, neighbourhood qualities, walkability)\\ Infer demographic characteristics from visual-spatial analysis (poverty, gentrification).\\ Understand non-physical (virtual) space and generic typologies\end{tabular} &  \\ \cmidrule(r){1-2}
\multicolumn{1}{@{}l|}{Benefits} & \begin{tabular}[c]{@{}l@{}}Automation of labour-intensive work\\ Methodological innovation for novel understanding of built environment based on existing data\\ New insights through transdisciplinary approaches\\ Leveraging existing datasets\end{tabular} &  \\ \cmidrule(r){1-2}
\multicolumn{1}{@{}l|}{Challenges} & \begin{tabular}[c]{@{}l@{}}Generalisability of findings\\ Reproducibility and accuracy of processes \\ Data problems: access, quality, reliability, interoperability, frequency\end{tabular} &  \\ \cmidrule(r){1-2}
\multicolumn{1}{@{}l|}{Future directions} & \begin{tabular}[c]{@{}l@{}}Transdisciplinary approaches to data curation for both qualitative and quantitative visual-spatial\\analysis\\ Interoperability of visual with other data types to understand and rate spatial relationships or\\characteristics (e.g. to influence decision-making processes, policy, property value estimation,\\building condition survey)\\ Integration of multiple feature analysis with different built environment scales (e.g. to predict\\urban and developmental trends, infer use and occupancy)\\ Democratisation of building-related data\end{tabular} &  \\ \cmidrule(r){1-2}
\multicolumn{2}{@{}l@{}}{Machine learning objectives} &  \\ \cmidrule(r){1-2}
\multicolumn{2}{@{}l@{}}{\begin{tabular}[c]{@{}l@{}}Test or compare the expediency of classification algorithms\\ Improve the performance and accuracy of image analysis or ML systems (e.g., image segmentation, object or texture detection,\\object extraction)\\ Test methods to infer information from multiple datasets\\ Identify methods to exploit existing architectural and urban image data\end{tabular}} &  \\ \cmidrule(r){1-2}
\end{tabular}
\end{table}

This review demonstrated that studies of the built environment adopting computer vision methods fall under two main categories. The first is studies that automate classification tasks mirroring established manual methods of visual analysis such as the interpretation of architectural or urban elements. Machine learning offers hereby the means to perform quantitative or qualitative analysis at scale. The second are studies that result in new insights through methodological innovations that utilise machine learning as a tool for data processing and analysis to raise novel questions in architectural and urban studies. Table~\ref{tab:summary} summarises current research trends and potential future directions. While this review excluded research using mainly generative machine learning models, non-visual data, or 3D datasets, these should be considered in the future when comparing research trends and their value to built environment studies. 

Future research directions with built environment applications call for more integration between disciplines. For example, automating existing visual survey methods using computer vision can only make labour-intensive tasks more affordable and data analysis more significant when the visual-spatial criteria are clearly defined and the data is reliable. Conversely, research focused on computer vision problems is unlikely to be relevant to built environment studies unless it is part of a larger research process and question.

Some of the future challenges particularly arise around data. Differences in data specification, accuracy, interoperability, creation, and access need to be resolved to create comparable and integrated datasets. Existing image datasets tend to lack context and metadata beyond GPS coordinates and crowd-sourced labels. In the few cases where metadata is available, the accuracy and reliability of such data was insufficiently tested. For instance, GPS coordinates of street-view panoramas may be too inaccurate for some applications and crowd-sourced data (such as Open Street Maps or Flickr) is prone to human errors or outdated. Meanwhile, elements of the built environment can be located and identified in vastly different ways across geographic information systems and administrations. Thus resolving data-related problems is especially important when automating the collection of large amounts of data, whether once or at regular intervals, which in the past was highly labour intensive. A more collaborative approach to data curation is also needed to ensure label accuracy.

Another key challenge is the reproducibility and comparability of computer vision-driven architectural research. While many works so far have been exploratory, future research is likely to seek to confirm current results and improve the state of the art. This requires more transparency in terms of both implementation and dataset availability, as well as the definition of a shared evaluation framework that speaks to both computer scientists and architects while staying connected to real-world applications.

A promising direction for future research is the integration of visual and non-visual data sources. Research might consider both building elements and context and combine image sources (aerial and street views) and other data such as statistical information (e.g. census data, household size, property type information, building age, and socio-economic or environmental statistics). The use of visual indicators in conjunction with existing building-related data, at a specific point but also over time, has significant potential to create new interdisciplinary approaches, improving upon both quantitative and qualitative research methods. This can in turn inform design decisions, building safety assessment and maintenance, real estate evaluation, and planning or socio-economic policies. 




\begin{acks}
This work is supported by the \grantsponsor{}{Prosit Philosophiae Foundation}{}.
\end{acks}

\bibliographystyle{ACM-Reference-Format}
\bibliography{bib/references}



\end{document}